\documentclass{IEEEcsmag}

\usepackage[colorlinks,urlcolor=blue,linkcolor=blue,citecolor=blue]{hyperref}

\usepackage{upmath}
\usepackage{graphicx}
\usepackage{amsmath}
\usepackage{subcaption}

\jvol{XX}
\jnum{XX}
\paper{8}
\jmonth{May/June}
\jname{Computer Graphics and Applications}
\pubyear{2021}

\begin{document}

\sptitle{Applications}
\editor{Editor: Name, xxxx@email}

\title{Predicting Surface Reflectance Properties of Outdoor Scenes Under Unknown Natural Illumination}

\author{Farhan Rahman Wasee}
\affil{Immersive and Creative Technologies Lab, Department of Computer Science, Concordia University}

\author{Alen Joy}
\affil{Immersive and Creative Technologies Lab, Department of Computer Science, Concordia University}

\author{Charalambos Poullis}
\affil{Immersive and Creative Technologies Lab, Department of Computer Science, Concordia University}

\markboth{Applications}{Predicting Surface Reflectance Properties of Outdoor Scenes Under Unknown Natural Illumination}

\begin{abstract}
Estimating and modelling the appearance of an object under outdoor illumination conditions is a complex process. Although there have been several studies on illumination estimation and relighting, very few of them focus on estimating the reflectance properties of outdoor objects and scenes. This paper addresses this problem and proposes a complete framework to predict surface reflectance properties of outdoor scenes under unknown natural illumination. Uniquely, we recast the problem into its two constituent components involving the BRDF incoming light and outgoing view directions: (i) surface points' radiance captured in the images, and outgoing view directions are aggregated and encoded into reflectance maps, and (ii) a neural network trained on reflectance maps of renders of a unit sphere under arbitrary light directions infers a low-parameter reflection model representing the reflectance properties at each surface in the scene. Our model is based on a combination of phenomenological and physics-based scattering models and can relight the scenes from novel viewpoints. We present experiments that show that rendering with the predicted reflectance properties results in a visually similar appearance to using textures that cannot otherwise be disentangled from the reflectance properties.
\end{abstract}

\maketitle

\chapterinitial{T}he creation of photo-realistic images is a well-studied yet non-trivial process which involves solving the rendering equation \cite{10.1145/15922.15902} at each surface point of the scene visible in a camera's image plane. In order to calculate it one has to provide (i) information about the geometry of the scene, (ii) the reflectance properties of the surface points in the scene, and (iii) information about the illumination conditions. Given this information as input, the solution provides an approximation of the outgoing radiance at each surface point which is then mapped onto a pixel on the camera's image plane to create the photo-realistic image. Indeed being a well-studied process, for the past few decades numerous techniques have been proposed primarily focusing on the main challenges of (i) physics-based reflection models, \cite{cook1982reflectance, he1991comprehensive} and (ii) efficient and effective processing of the input information e.g. sampling \cite{cook1986stochastic}, photon mapping \cite{jensen1996global, jensen2001realistic}, radiosity \cite{cohen1988progressive, heckbert1990adaptive}, path tracing \cite{lafortune1993bi}.

Inverting the aforementioned process is itself a complex, non-trivial process known as Inverse Global Illumination (IGI). The objective is to recover the reflectance properties of the surface points in the scene given (i) a set of images capturing the scene, information about the geometry of the scene, and (ii) the illumination conditions at the time the images were captured. Recovering the reflectance properties of the scene then allows its relighting under any arbitrary illumination conditions which has proven to be of enormous importance in the film industry. Research has shown that IGI is applicable in controlled environments e.g. where the geometry of the scene is captured with a laser scanner and the lighting conditions are recorded using a light-probe \cite{debevec2004digitizing, debevec2004estimating} in-sync with the image capture. 

In contrast, little work has been done on IGI for large-scale scenes such as urban areas with complex illumination conditions. This is expected, considering that capturing the geometry of large-scale areas with a high-enough accuracy is a very expensive and time-consuming process. To address this problem, a number of techniques have been proposed in the past few years using deep neural networks e.g. \cite{rematas2016deep, chen2019learning, kato2018neural} in which given images of the scene and in some cases additional information extracted from those images e.g. depth map, semantic maps \cite{meshry2019neural} the network produces renders of the scene from novel viewpoints. The results are very impressive however the user has no control over the output of the network i.e. the reflectance properties are embedded in the network and cannot be extracted. This results in images of the scene which may not be (and most of the times are not) consistent between different novel views.

In this paper we address the problem of recovering the reflectance properties of objects from images. Uniquely we recast the problem of recovering the bidirectional reflectance distribution function (BRDF) of four parameters $f_{r}(\omega_{i}, \omega_{o})$ into its two constituent components. In the first component, all samples of the reflected radiance captured in the images - from arbitrary view directions - are aggregated on a per-triangle basis. Assuming a single directional light source in the scene, each surface point is represented as a unit sphere and information about the outgoing view directions $\omega_{o}$ and radiance of the surface points contained in each surface triangle are encoded in reflectance maps.

In the second component, a deep neural network is used to reduce the parameter space of the BRDF from the 4-parameters relating to the incoming light direction $\omega_{i}$ and outgoing view direction $\omega_{o}$, to only the 2-parameters relating to $\omega{o}$. The network is trained on a vast number of reflectance maps corresponding to renders of the unit sphere with different BRDFs being lit from arbitrary light source directions rendered from a fixed viewpoint, leading to embedding information about arbitrary incoming light directions $\omega_{i}$. Thus, the network learns a function for mapping a reflectance map corresponding to a render of a unit sphere being lit from an arbitrary light direction to a 7-vector representing the reflectance properties of the unit sphere. Comparative analysis between various network architectures show that residual neural networks outperform alternative architectures.


To summarize, the contribution of this work is a complete two-stage framework for recovering reflectance properties of objects in scenes from a sparse set of images. The proposed approach has the significant advantage of making the problem tractable for large objects/scenes by not requiring the direct measurement of the BRDFs. Uniquely, we recast the problem into two sub-problems involving the BRDF's incoming light and outgoing view directions, $\omega_{i}$ and $\omega_{o}$, respectively: first by aggregating information on surface outgoing radiance and outgoing view directions $\omega_{o}$ into reflectance maps, and then by estimating the reflectance properties of a surface given sparse reflectance maps with a network embedding information about arbitrary incoming light directions $\omega_{i}$. Moreover, in contrast to recent work, recovering the reflectance properties of each surface in the scene parameterized as a 7-vector, gives full-control when relighting the scene under novel lighting conditions.

\vspace{-10pt}
\section{RELATED WORK}
\label{sec:related_work}

Appearance capture and modeling, and inverse rendering in particular have been active research topics for well over four decades. During this time large number of techniques have been proposed to address the challenging problem of accurate estimation of reflectance properties from images. Below, we provide a brief overview of the most recent and relevant state of the art categorized according to the line of research followed i.e. procedural-based algorithms, and deep-learning-based.

\vspace{-10pt}
\subsection{PROCEDURAL-BASED}
Traditional systems need human assistance, a controlled environment for setting up experiments that capture the appearance of a scene. To capture the illumination or environment, various instruments have to be used. For instance, a laser scanner that can capture the scene geometry, light probes that are used to capture the environment, the light information, etc. \cite{marschner1997inverse} presents one such attempt to estimate the effects of light in an image with the help of the scene geometry and camera parameters. The system tries to solve a least square system to find the light distribution which is then used to compute the changes required to go to the target lighting conditions. Another significant work in reflectance estimation was done by Yu et al.\cite{yu1999inverse}. It attempts to recover diffuse and specular reflectance, and model it using a low parameter reflectance model. This requires several calibrated images taken under predefined lighting conditions alongside the geometric model of that scene. Then they try to optimize for the reflectance parameters iteratively. Romeiro et al.\cite{romeiro2008passive} were also one of the first to show how reflectance properties can be estimated under known shape and illumination. 

Debevec et al \cite{debevec2004digitizing, debevec2004estimating} proposed a technique for estimating reflectance properties of a landmark under natural illumination. The technique involves scanning the scene geometry using a laser scanner, taking multi-view photographs from different locations, and also capturing the illumination condition for each of the images. In addition to all of these, a subset of BRDF measurements is also taken from different surfaces amongst the scene. Afterwards, an iterative global illumination procedure is performed that estimates the reflectance properties and re-renders the scene. The images are then compared with the original ones and the parameters are tuned to minimize the errors. Lombardi and Nishino \cite{lombardi2012reflectance} reformulated the problem of acquiring reflectance from images using a probabilistic approach. Despite adding valuable contribution to the field, the downside with many of these approaches is they rely heavily on the setup of the scene, and will need a different approach if one or more information is not available about the scene. Laser scanning a scene is not easy, and it is not always practical to design a method for which the geometry, images, BRDFs, etc will be readily available to estimate the materials. Also, a few of these experiments work with simplistic objects i.e. images of spheres against a solid background or an environment map. Hence, they do not scale up with complex geometric shapes.

\vspace{-10pt}
Barron and Malik \cite{barron2012color, barron2012shape} proposed two separate techniques for recovering shape, diffuse albedo and illumination from an image. The first of these two techniques focused on achromatic images while the latter worked on colored images. However, the problem with this approach was the surface materials were modeled using the Lambertian reflection model. Glossiness or the concept of roughness is not defined in the Lambertian model. Which makes the technique non physics based and thus it is limited in its ability when it comes to real-world materials. Brelstaff and Blake\cite{brelstaff1987computing} proposed a method that attempted to address this issue of identifying specularity in images. Their method works well but it can only identify specular shiny regions from images. Abe et al.\cite{abe2012recognizing} use several feature detectors to extract features from images and use an SVM to rank the images based on a few predefined attributes, which are glossiness, roughness, transparency and coldness. Goldman et al. \cite{goldman2009shape} estimated per-pixel mixture weights and a combination of parametric materials. However, this approach requires a few HDR input images which were captured under known lighting conditions, which is again difficult to obtain.

Over the years, several different techniques have been proposed to obtain material properties from images. However, using traditional techniques most often work by relying on laboratory setup. Moreover, some of these techniques pose constraints on additional factors like illumination, shape, etc. Shapes of objects can be of countless types, illumination can have a lot of variations. Besides, materials can have subtle variations producing changes in appearance - addressing all of these factors at once is difficult using traditional techniques and rule-based approaches. Therefore, estimating realistic materials properties under varying illumination, i.e estimating diffuse, specular, roughness components can be challenging with these methodologies. 


\vspace{-10pt}
\subsection{DEEP-LEARNING-BASED}
The revolution of deep learning as seen from the very early results from has opened a new paradigm of ways to tacking the inverse problem of relighting and material estimation, especially with the development of various deep learning frameworks \cite{NIPS2019_9015,abadi2016tensorflow,jia2014caffe}. Lately differentiable rendering has been used to solve similar subtasks as seen in \cite{chen2019learning,Li_2020_CVPR,Liu_2017_ICCV,Azinovic_2019_CVPR,Yu_2019_CVPR,nguyen2018rendernet}.

Dror er al. \cite{dror2001estimating} estimate reflectance properties from just 6 predefined types specified by the Ward model\cite{ward1992measuring} and trained a classifier using synthetically created images of spheres.

Rematas et al. \cite{rematas2016deep} addressed this inverse problem by creating reflectance maps from specular materials in natural lighting conditions. Their technique assumes that the object is made up of a single material that is consistent throughout its geometry. They proposed two modules in their system, one of which directly estimates the reflectance maps from the image while the other first predicts per-pixel surface normals that are used to compute a sparse reflectance map. It is then used in interpolation to create a denser map.

Meka et al. \cite{meka2018lime} used a system with 5 smaller deep networks to estimate materials in order to perform material transfer. An image is taken as input which is first masked from its background and then further processed to estimate diffuse and specular components of the material. The system works with objects with simpler shapes and estimate a single material for each image, and has a few limitations like not supporting global illumination and color specularity.

Kim et al. \cite{kim2017lightweight} built a lightweight system to estimate surface reflectance from images in real-time. They estimate the surface reflectance by using a simple BRDF representation and by estimating surface albedo and gloss instead of estimating the full 4D BRDF function. They use two networks to estimate the BRDF which enables rendering in illumination and view independent conditions.

Philip et al. \cite{philip2019multi} proposed an end to end approach for relighting outdoor scenes, mostly of landmarks. Their system takes a set of multi-view images, compute a proxy 3D geometry. A part of their system focus on the shadow removal and refining process using another sub network. There system has also been trained on photo-realistic synthetic data generated using physics based ray-tracing. A similar approach has been taken by Meshry et al. \cite{meshry2019neural} also attempt to do a total scene capture which takes publicly available images of landmarks as inputs and use off the shelf reconstruction algorithms to retrieve the geometry. 

Chen et al. \cite{chen2019learning} uses one of the recent approaches of using a neural renderer to tackle similar sub-problems. In there work, they designed a neural renderer which attempts to synthesize novel viewpoints and re-illumination by defining the appearance as a combination of environment lighting, object intrinsic attributes, and the light transport function (LTF) which are learnable through training. The scenes are mainly of small scale objects and it does not work particularly well in the presence of highly specular objects.

The authors in \cite{Murmann_2019_ICCV} published a dataset of indoor scenes composed of a fixed number of predefined material categories in multi-illumination conditions. Along with the dataset they tackled the problem of scene illumination prediction by predicting the light probe for the given input image. Also they formulated image relighting as an image to image translation problem where they train a network to learn the direct mapping (without factoring out the material properties) between two images  taken from the same viewpoint under two lighting conditions. The system does not estimate reflectance properties throughout the process and is limited to close shot images and that are taken indoors.


Other methods \cite{song2019neural,gardner2017learning,Hold-Geoffroy_2017_CVPR} capture the illumination conditions, or the environment that the image was taken, to perform manipulations such as inserting a new object into the scene, etc. However, these methods do not give a way to reintroduce new lighting conditions and re-rendering. 

To summarize, despite all the recent advancements and contributions in the field of inverse rendering and material estimation the procedural methods generally work well under constrained environment, controlled lighting and/or with objects with simpler geometry. On the other hand, the recent deep learning techniques produce results that provide limited or no control from the end user. Also, most often, the internal representation of what is going on under the hood in these relighting models is very hard or nearly impossible to interpret. Some of these techniques take in images, and re-render using different lighting condition. Most often, the user has no knowledge on how the network is internally registering the materials. Whereas, in our technique, the materials are well defined, and the viewpoints can be controlled. Each of the objects in the scene has specific reflectance properties assigned to the geometry. Besides, the material reflectance model used in this work is physically based and can be used to model any realistic material.

\begin{figure}[!ht]
	\centering
	\includegraphics[width=0.5\textwidth]{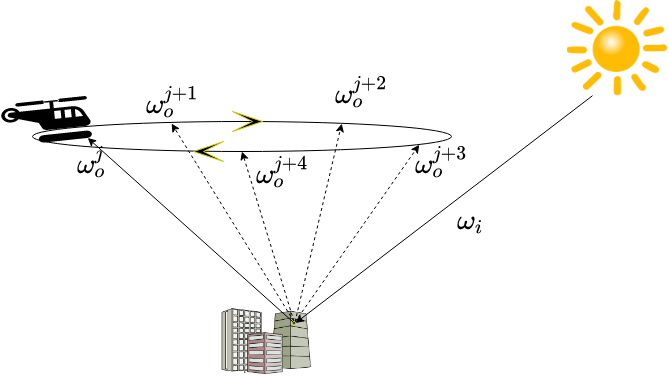}
	\caption{The capture process of wide-area motion imagery is performed during the day over a few minutes. The aerial sensor is attached to an aircraft that is orbiting the scene. Each image $j$ captures the reflected radiance \textbf{$\omega_{o}^{j}$} from all visible surface points in the scene. The sun serves as the single directional light source illuminating the scene from direction $\omega_{i}$ and is considered to be fixed throughout the process.}
	\label{fig:capturing_process}
\end{figure}

\vspace{-10pt}
\section{REFLECTION MODELS}
\label{sec:reflection_models}
The reflectance at a surface point $P$ is modeled by the bidirectional reflectance distribution function (BRDF) and is given by,
\begin{equation}
	f_{r}(P, \omega_{i}, \omega{o}) = \frac{L(P, \omega_{o})}{L(P, -\omega_{i})cos(\phi)m(\Omega)}
	\label{eq:brdf}
\end{equation}
where $\Omega$ is the solid angle subtended by the light source at the surface point $P$, $m(\Omega)$ indicates its measure, $\omega_{i}$ and $\omega_{o}$ are the incoming (from light source) and outgoing (to viewpoint) directions, and $\phi$ is the colatitude of the light source. The parameters $\omega_{i}$ and $\omega_{o}$ for the incoming and outgoing directions are further parameterized by the azimuth angle $\phi$ and zenith angle $\theta$ making the BRDF a function of four parameters $f_{r}(\phi_{i}, \theta_{i}, \phi_{o}, \theta_{o})$.

\textbf{Acquisition.} The acquisition of the BRDF (Eq. \ref{eq:brdf}) at a surface point $P$ involves measuring the amount of the incoming radiance that is reflected at $P$ for all combinations of $(\omega_{i} \in \textbf{S}^{2}_{-}, \omega_{o} \in \textbf{S}^{2}_{+}$. We use $\textbf{S}^{2}$ to denote the unit sphere in 3-space i.e. the set of all possible directions in which light can flow. $\textbf{S}^{2}_{+}$ consists of all vectors pointing away from the surface, while $\textbf{S}^{2}_{-}$ consists of vectors pointing into the surface. 

Acquiring BRDF measurements under lab conditions is a straight-forward process involving the use of a gonioreflectometer located in a dark room. Having no additional light sources avoids interference with the measurements. A spotlight illuminating a sample placed at the center of the sphere is mounted on a spherical arm. Its position is varied and a sensor captures the amount of light bouncing off the sample. 

For large objects located outdoors, acquiring BRDF measurements is an extremely difficult and nontrivial task primarily because of the complex illumination and the fact that it cannot be controlled. During the day the sun serves as the only light source of the scene whereas at night a number of light sources (e.g. light posts, moon, flashlights, etc) may contribute to the illumination of the scene. In either case controlling the lighting conditions is almost impossible and unless strong assumptions are made about the scene acquiring BRDF measurements for all surface points of the scene becomes intractable. 

Perhaps the most relevant work on acquisition of BRDFs of outdoor objects under complex illumination condition was the work of Debevec et al in \cite{debevec2004estimating}. In their work, the authors had ground-access to the structure during the night which allowed them to capture BRDF measurements using a custom-made gonioreflectometer-like device from four representative areas of the structure. Based on the assumption that the entire structure could be represented by those 4 BRDFs they recovered per-surface-point BRDFs and relit the structure under novel lighting and produce very realistic renders.

In our case  there are additional restrictions which exacerbate the difficulties of acquiring BRDFs: (i) Although the structure is outdoors we do not have ground-access to perform close-up experiments, (ii) The wide-area motion imagery is captured from an aerial sensor which is orbiting the scene, (iii) Due to the weak-perspective (i.e. the distance of the sensor from the scene is considerably larger than the range of depth within the scene) each image captures a large physical area, (iv) The image capture is done during the day. For these reasons, we are limited to image-based BRDF acquisition which is often incomplete and noisy. Figure \ref{fig:capturing_process} shows the capturing process.

\textbf{Representation.} Assuming all the problems relating to measuring BRDFs are overcome, there is still the issue of storage and representation. Using dense/full BRDF measurements can provide the most accurate results. The BRDF for a particular pair $(\omega_{i}, \omega_{o})$ can be retrieved or interpolated from nearby samples. This however increases computational complexity and imposes a significant cost for performing effective sampling during rendering. In this work we use a combination of phenomenological and physically-based scattering models. 

A diffuse surface scatters incident illumination equally in all directions. These type of surfaces are modeled as a Lambertian reflection model. This is a phenomenological model which assumes that the BRDF is constant and is given by $f_r^{L}(P, \omega_{i}, \omega_{o}) = \frac{\rho}{\pi}$ where $\rho$ is the albedo i.e. the fraction of the arriving light energy that is scattered. The reflected radiance $L_{o}$ varies linearly with the incident radiance $L_{i}$ and is given by $L_{o} = f_{r}^{L}(P, \omega_{i}, \omega_{o})L_{i} = \frac{\rho L_{i}}{\pi}$.


The Lambertian reflection model does not account for cast shadows or specularities. Specular scattering is modeled using the micro-faceted Torrance-Sparrow BRDF \cite{torrance1967theory} given by,
\begin{equation}
	f_{r}^{TS}(P, \omega_{o}, \omega{i}) = \frac{D(\omega_{h})G(\omega_{o}, \omega{i})F_{r}(\omega_{o})}{4cos(\theta_{o})cos(\theta_{i})}
\end{equation}
where $D(.)$ is the Trowbridge and Reitz micro-facet distribution function \cite{trowbridge1975average}, $G(.)$ is the geometric attenuation term, $F_{r}(.)$ is the Fresnel function, and $\theta_{o}$ and $\theta_{i}$ are the angles between the normal at surface point $P$ and the outgoing direction $\omega_{o}$ and incoming direction $\omega_{i}$ respectively. The relation between the reflected radiance $L_{o}$ and the incoming radiance $L_{i}$ is then given by,
\begin{align}
	L_{o} = f_{r}^{TS}(P, \omega_{i}, \omega_{o}) L_{i} cos(\theta_{i}) d\omega_{i} = \\ L_{i} \frac{D(\omega_{h})G(\omega_{o}, \omega{i})F_{r}(\omega_{o})}{4cos(\theta_{o})cos(\theta_{i})} cos(\theta_{i}) d\omega_{i}
\end{align}

\vspace{-10pt}
\section{METHOD}
\label{sec:method}
In our work the acquisition of the wide-area motion imagery occurs during the day over a few minutes. The sun serves as the single directional light source illuminating the scene and is considered to remain fixed throughout the capturing process (Figure \ref{fig:capturing_process}). Typically, measuring the sun's direction and intensity requires additional captures using specialized equipment \cite{debevec2012single}. Although this may be appropriate when ground access to the scene is possible, this is not applicable in cases where the capture is performed using only aerial sensors. For this reason, we avoid the explicit measurement of light direction $\omega_{i}$ by training a neural network to predict the reflectance parameters given samples of the reflected radiance of an object under arbitrary directional illumination conditions. Formally, the neural network $\Phi(.)$ learns the mapping function $\Phi: I_{\mu} \longrightarrow \mu$ where $\mu$ is a low-parameter(7-vector) reflection model approximating the reflectance properties. The reflectance properties $\mu$ are defined in terms of the diffuse $k_{d}^{3}$ and specular $k_{s}^{3}$ components, and the surface roughness $r$. The fact that the incident radiance is arriving from a single fixed direction $\omega_{i}$ facilitates the representation of the BRDF of a surface point $P$ as a unit sphere $\textbf{S}^{2}$ where the reflected radiance value is stored at each point on the sphere's surface corresponding to each outward direction $\omega_{o} \in \textbf{S}^{2}_{+}$.

\begin{figure}[!ht]
	\centering
	\includegraphics[width=0.5\textwidth]{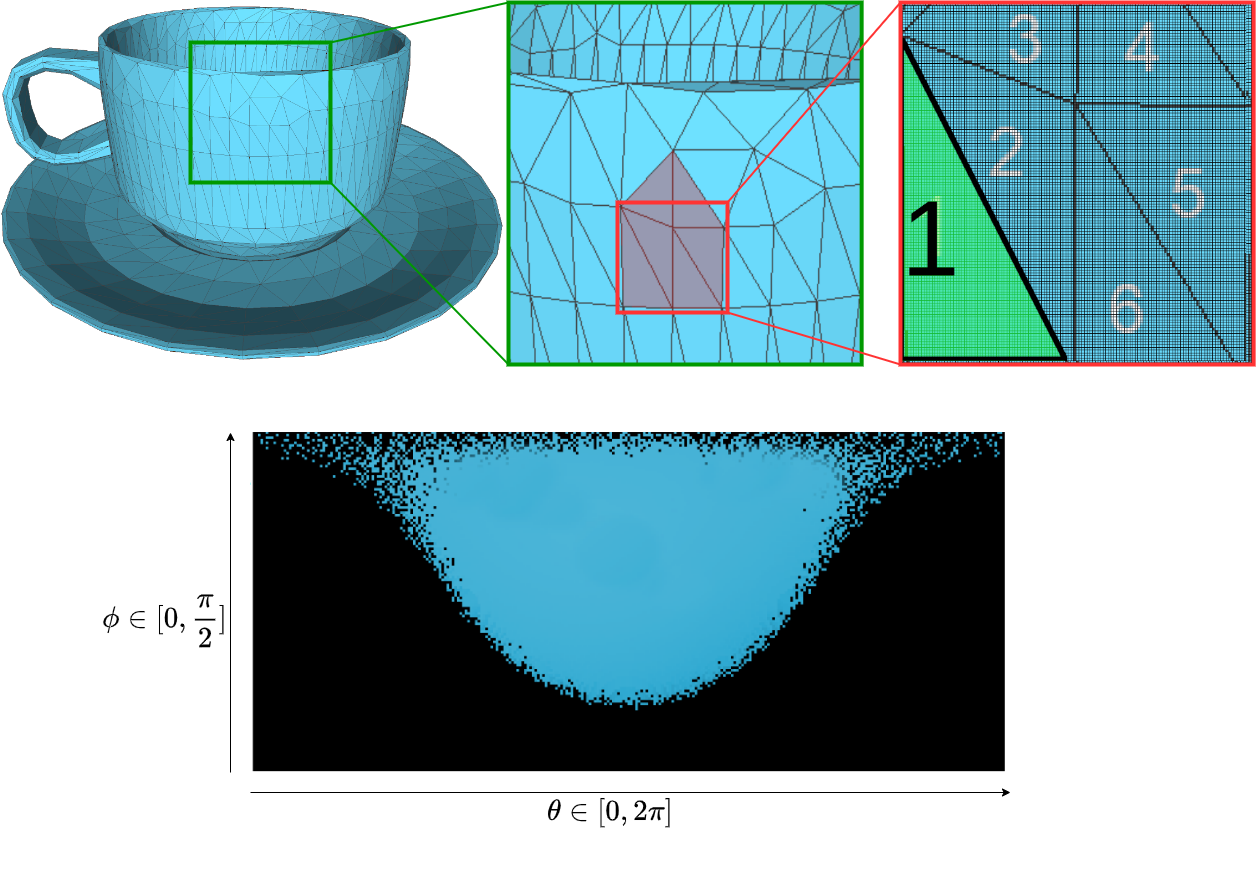}
	\caption{The top row shows sequential close-ups of the areas marked by the green and red squares. The bottom row shows the reflectance map produced by unwrapping the texture of a unit sphere used to store the aggregated reflected radiance towards each outgoing direction from all surface points contained inside the triangle numbered with 1.}
	\label{fig:example1}
\end{figure}

We assume that two surface points belonging to the same surface-triangle do not have a large difference between the angles of their outgoing viewing direction $\omega_{o}$ and the light direction $\omega_{i}$, and we aggregate the reflected radiance values on a per-surface-triangle basis. This assumption is based on (i) the fact that the incoming light direction and surface normal are identical for all surface points belonging to a surface-triangle, (ii) the weak-perspective nature of the capturing process i.e. the light source is fixed during the capture process and at an infinite distance from the scene when compared to the radius of the sphere, and (iii) surface points belonging to the same surface-triangle almost always have the same reflectance properties. Therefore, the reflected radiance values of each surface point contained in the same surface-triangle are aggregated: for each surface point, we calculate its spherical coordinates in terms of $(r, \theta, \phi)$ where the $r$ is the distance of the point from the origin of the sphere representing it, whereas $\theta$ and $\phi$ are the azimuth and altitude for that point. Using this parameterization, every point of the sphere adds a contribution to the reflectance map. The unwrapped texture of each sphere forms the reflectance map for that surface-triangle. Figure \ref{fig:example1} shows a simple example of this process. The top row shows sequential close-ups of the areas marked by the green and red squares. In this example, the bottom row shows the reflectance map produced by unwrapping the texture of a unit sphere used to store the aggregated reflected radiance towards each outgoing direction from all surface points contained inside the triangle numbered with 1. Thus, the input to the network $I_{\mu}$ is the reflectance map formed by unwrapping the texture of the unit sphere used to represent the reflected radiance towards each outgoing direction $\omega_{o} \in \textbf{S}^{2}_{-}$.

\vspace{-10pt}
\subsection{NEURAL NETWORK}
\label{subsec:neural_network}

\subsubsection{Training data}
\label{subsubsec:training_data}

A physics-based renderer \cite{pharr2016physically} was employed for generating the training dataset. In order to ensure variability in the training dataset the following conditions were varied: (i) reflectance properties (diffuse, specular, roughness), (ii) camera viewpoint. The scene consisted of a unit sphere with uniform, isotropic, and opaque reflectance properties, a perspective camera, and a single fixed directional light source. 

\begin{figure}[!ht]
	\centering
	\includegraphics[width=0.5\textwidth]{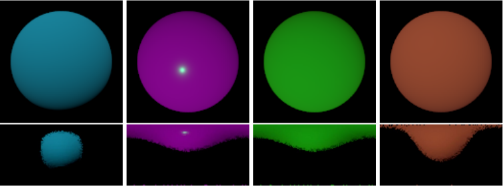}
	\caption{Top row: Spheres rendered with different BRDFs with varying diffuse, specular, and roughness parameters, and different viewpoints. Bottom row: The reflectance maps formed by unwrapping each sphere.}
	\label{fig:reflectance_maps}
\end{figure}

The camera orientation is defined in spherical coordinates $(\theta_{c}, \phi_{c})$ with $\theta \in \{\frac{\pi}{9}, \frac{5\pi}{18}, \frac{4\pi}{9} \}$ and $\phi \in [0,2\pi]$ at intervals of $\frac{\pi}{4}$. The reflectance properties are defined in terms of the diffuse and specular components given by $[ k_{d}^{r}, k_{d}^{g}, k_{d}^{b}, k_{s}^{r}, k_{s}^{g}, k_{s}^{b}, r ]$ where $k_{d}^{r}, k_{d}^{g}, k_{d}^{b}$ is the diffuse component, $k_{s}^{r}, k_{s}^{g}, k_{s}^{b}$ is the specular component, and $r$ is the surface roughness. These parameters are set to different values in their respective range creating a total of 24,000 unique materials. The parameters are set according to the following two configurations: 20,480 of the materials are uniformly sampled and 3,520 materials are randomly sampled. Each of the materials were rendered from 24 different view points which resulted in a total number of 576,000 images.   Due to the reciprocal relation between the viewpoint and the light direction, the latter remains fixed during the creation of the renders. In this work we use a directional light source with a constant white color. Figure \ref{fig:reflectance_maps} shows examples of renders from the training dataset. The bottom row shows reflectance maps corresponding to the spheres in the top row, each with a different BRDF. 

Typically, the resolution of the reflectance map is $\pi \times 2\pi$. Our experiments have shown that even for smaller resolutions the accuracy of the network's predictions remains unaffected while training time significantly reduces. For this reason, we opt for a smaller resolution of the reflectance maps i.e. $\frac{\pi}{3} \times \frac{2 \pi}{3}$.


\vspace{-20pt}
\section{NETWORK}
\label{sec:network}

\subsection{NETWORK ARCHITECTURE}
A Wide ResNet-50 \cite{zagoruyko2016wide} architecture was chosen for training. The input is a reflectance map of size $60\times120\times3$ and the output is a low-parameter reflection model represented as a $7\times1$ vector. A sigmoid activation function is used at the end after the fully connected layers to constrain the range of values of the 7-vector to $[0-1]$. Data augmentations were employed to enhance the performance and improve robustness to variations during inference. We used a 80/20 split for training and validation. An additional 30,000 images were generated and retained for subsequent testing. The initial learning rate was set to 0.0003 and was gradually reduced by 10\% after every 20 epochs. 

\vspace{-10pt}
\subsection{LOSS FUNCTION}
For the training of the network, the loss function is a weighted sum-of-squared-errors. Except from the roughness, the other two parameters are 3-vectors ($k_{d},k_{s}$). Therefore, we set the roughness' weight $\theta_{1}$ to three such that the importance equals that of $k_{d},k_{s}$. Furthermore, our experiments have shown that the uniform sampling of the parameters can cause a strong bias against $k_{s}$ in the generated training data. Materials with high roughness produce almost identical results regardless of the value of $k_{s}$. Hence, the weight $\theta_{2}$ is introduced that causes the network to down-weigh the importance of the $SSE_{k_{s}}$ whenever the roughness $r$ is high. The loss function $E(k_{d},k_{s}, r)$ is given by,
\begin{equation}
	\label{equ-loss}
	E(k_{d},k_{s}, r) = SSE_{k_{d}} + \theta_{1} \times SSE_{k_{s}} + \theta_{2} \times SSE_{r} 
\end{equation}
where $\theta_{1} = (1-\sqrt[3]{r})$, $\theta_{2} = 3$, and $SSE$ is the sum-of-squared-errors. 

\vspace{-10pt}
\subsection{DATA AUGMENTATIONS}
Data augmentation was introduced during the training process to increase the variability and to simulate the sparsity exhibited in the reflectance maps. The maximum number of samples a point can have is equal to the number of images capturing it. However, due to the nature of the data acquisition process for multi-view imagery, it is common to have incomplete data for a surface point due to occlusions or invisibility from the cameras. This, in practice, results in a very low number of samples making the reflectance maps very sparse. To address this, as part of the data augmentation process, we randomly scatter salt-and-pepper noise, localized random noise on the RGB pixel values, and masking out. Additionally, random vertical and horizontal flips were applied to the reflectance maps to increase variability -in view and light directions- and avoid over-fitting. We opt out of using affine transformations e.g. scaling, shearing, etc, as data augmentations. These would cause significant changes in the appearance of the image leading in problems relating the appearance of the object with respect to the surface normals, light direction, and viewpoint direction. 

\vspace{-10pt}
\subsection{TRAINING AND INFERENCE}
\label{subsec:training_and_inference}
\textbf{Training.} The Wide-Resnet50 architecture was trained for 90 epochs. Using a step learning rate with an initial learning rate of $0.0003$, and a reduction of 10\% at every 20 epochs, resulted in lower validation loss and a smooth training loss. The network was trained on an NVidia V100 GPU. Instead of using dropout as a regularization technique, our system relies on the data augmentations and weight decay to avoid over-fitting on the training set. We used a batch size of 256. In addition to performing on the data augmentations described above, we also normalize the data during training, validation and testing to have a mean of 0 and a standard deviation of 1.

\noindent
\textbf{Inference.} For testing, our system requires multi-view images and associated camera poses, and the geometry of the scene. In cases where the geometry was not available, we used COLMAP\cite{schoenberger2016mvs, schoenberger2016sfm}. At the first stage of our system, a buffer is created for all the images using the camera poses and the 3D model. Lets denote the $n$ number of multi-view images as $I = \{I_{1}, I_{2}, I_{3}, ...,I_{n}\}$ and the camera poses as $C=\{C_{1}, C_{2}, C_{3}, ...,C_{n}\}$, then the system would generate buffers $B=\{B_{1}, B_{2}, B_{3}, ...,B_{n}\}$ where $height(I_{i})=height(B_{i})$, $width(I_{i})=width(B_{i})$ and $i=1,2,...,n$.

\vspace{-10pt}
The buffers record a few pieces of information about the closest intersection for each of the pixels. These information include, \textbf{(i.)} ID of the closest intersecting triangle from the mesh, denoted by $(t_{id})$ \textbf{(ii.)} World coordinate of the intersecting point $(p_{x}, p_{y}, p_{z})$ \textbf{(iii.)} Normal of the intersecting point $(n_{x}, n_{y}, n_{z})$ \textbf{(iv.)} Projected point in pixel/image space $(u, v)$. Therefore,
$(t_{id}), (p_{x}, p_{y}, p_{z}), (n_{x}, n_{y}, n_{z}), (u, v) \in B_{1},...,B_{n}$.

\vspace{-7pt}
After calculating the buffers, the system proceeds by reading in a set of $({I_{i}, B_{i}, C_{i}})$. Then for each $(u, v)$, it collects the RGB value from $I_{i}$. Then it has to find the index to place this RGB value in the reflectance map associated with $t_{id}$. The system calculates the index by computing the angle between the normal $(n_{x}, n_{y}, n_{z})$ and using the viewpoint direction from $C_{i}$. The resultant angle $(\theta,\phi)$ in spherical coordinates is used to place the extracted RGB value into the reflectance map associated with $t_{id}$. This step is repeated for all the pixels in $I_{i}$. After all sets of $({I_{i}, B_{i}, C_{i}})$ are processed, all the triangular surfaces in the scene will have their reflectance maps computed. 

\vspace{-5pt}
In practice, a triangle $t_{id}$ in very small compared to the size of the entire model. Therefore, the visible part of any triangle $t_{id}$ in an image ${I_{i}}$ might not have a lot of $(u, v)$ recorded against itself in the buffer. This results in making the reflectance maps very sparse. The number of images that are available for a test scene will also impact the number of samples. The reflectance maps for each of the triangles go into the network for prediction. The output of the network is then assigned to each of the triangles individually. A slight error in the estimation can cause the re-rendered image to appear triangulated.

\section{EXPERIMENTAL RESULTS}
\label{sec:experiments}
We conducted a series of experiments for the evaluation of the proposed approach. In the first part of this section, we present experiments on different network architectures. In the second part, we present the results on synthetic images and several outdoor scenes.

\vspace{-10pt}
\subsection{OPTICAL NETWORK ARCHITECTURE}
\label{subsec:network_comparison}
To determine the optimal network architecture, we evaluated eight neural networks' performance on a 30,000 test set. We experimented on the different variants of the ResNet architecture\cite{he2016deep}. To experiment on deeper and wider networks we have used multiple variants of ResNext\cite{xie2017aggregated} and WideResNets \cite{zagoruyko2016wide}. AlexNet \cite{krizhevsky2012imagenet} was chosen as a baseline to show how a model with fewer parameters and less depth fits our data. The most recent architecture and state-of-the-art network that we have used is the EfficientNet \cite{tan2019efficientnet}. 

The reflectance maps in the test set have variations in viewpoints and materials. The network predictions were used to re-render the spheres under original lighting and from the same viewpoint. To analyze each of the networks' performance, we measured the error in the parameters and the error in image-to-image comparisons. We report our results on three metrics: (i) Peak Signal to Noise Ratio (PSNR), (ii) Normalized Root Mean Squared Error (NRMSE), and (iii) Structural Dissimilarity Index (DSSIM) \cite{wang2003multiscale} which unlike RMSE or PSNR which only calculate the errors amongst pixels, it utilizes structural information by assuming that pixels have strong inter-dependencies when they are spatially close within a neighborhood. Table \ref{table:comparison} shows the results. As it is evident the WideResnet-50 outperforms all others in all the three measures.

\begin{table}
		\hspace{-20pt}
		\begin{tabular}{ |p{3cm}||p{1.25cm}|p{1.25cm}|p{1.25cm}|  }
			\hline
			Architecture & PSNR & NRMSE & DSSIM\\
			\hline
			
			AlexNet\cite{krizhevsky2012imagenet} & 41.6841 & 0.0564 & 0.00676\\
			\hline
			ResNet-18\cite{he2016deep} & 45.4900 & 0.0354 & 0.00391\\
			\hline
			ResNet-34\cite{he2016deep} & 46.3642 & 0.0349 & 0.00401\\
			\hline
			ResNet-50\cite{he2016deep} & 47.2551 & 0.0325 & 0.00398\\
			\hline
			Wide Resnet-50\cite{zagoruyko2016wide} & \textbf{48.0526} & \textbf{0.0287} & \textbf{0.00326}\\
			\hline
			ResNext-50\cite{xie2017aggregated} & 47.4328 & 0.0309 & 0.00375\\
			\hline
			ResNext-101\cite{xie2017aggregated} & 47.5832 & 0.0307 & 0.00357\\
			\hline
			EfficientNet B0\cite{tan2019efficientnet} & 47.7031 & 0.0306 & 0.00358 \\
			\hline
		\end{tabular}
	\caption{Results on the test set across different model architectures using PSNR, NRMSE and DSSIM as metrics. The test set had 30,000 samples.}
	\label{table:comparison}
\end{table}

\vspace{-10pt}
\subsection{SYNTHETIC IMAGES}
\begin{figure}[!ht]
	\centering
	\begin{subfigure}[t]{0.15\textwidth}
		\includegraphics[width=\textwidth]{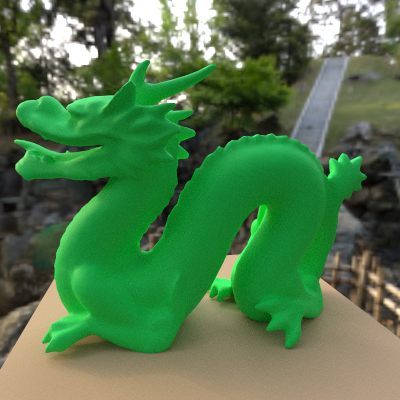}
		\caption{}
		\label{fig:dragon_gt}
	\end{subfigure}
	\begin{subfigure}[t]{0.15\textwidth}
		\includegraphics[width=\textwidth]{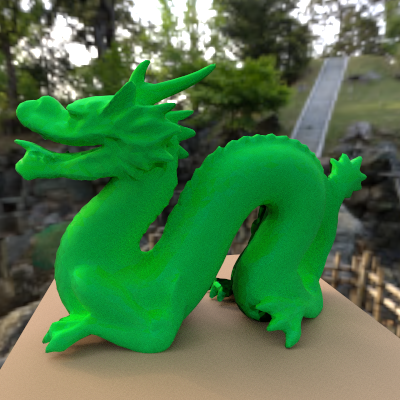}
		\caption{}
		\label{fig:dragon_pred}
	\end{subfigure}
	\begin{subfigure}[t]{0.15\textwidth}
		\includegraphics[width=\textwidth]{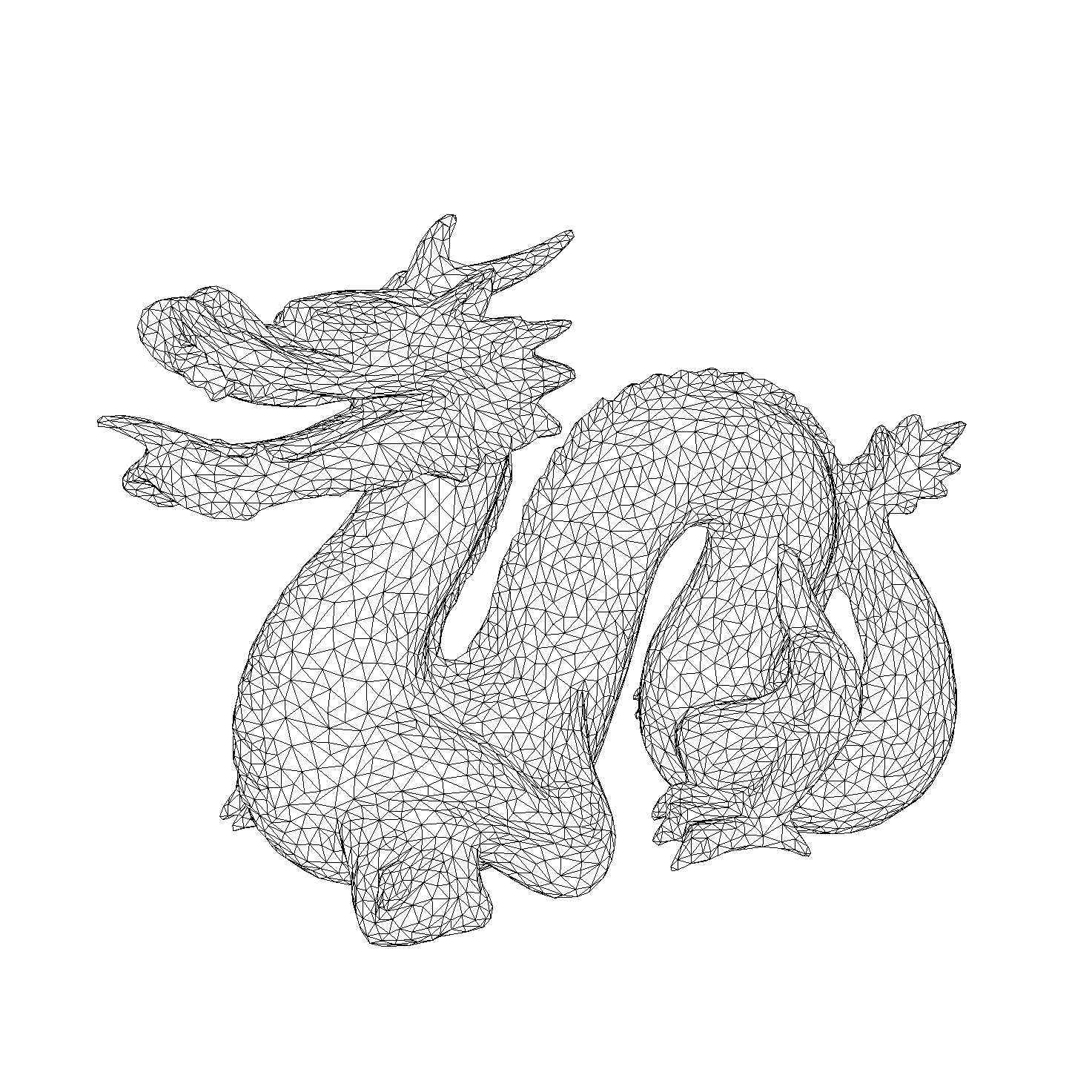}
		\caption{}
		\label{fig:dragon_mesh}
	\end{subfigure}
	
	\begin{subfigure}[t]{0.15\textwidth}
		\includegraphics[width=\textwidth]{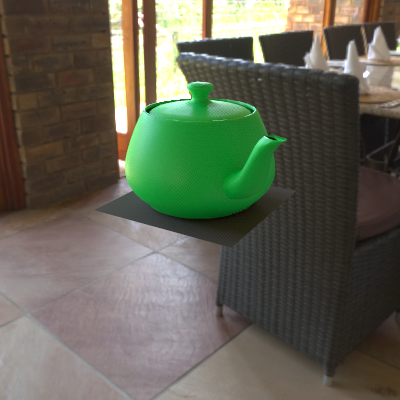}
		\caption{}
		\label{fig:teapot_gt}
	\end{subfigure}
	\begin{subfigure}[t]{0.15\textwidth}
		\includegraphics[width=\textwidth]{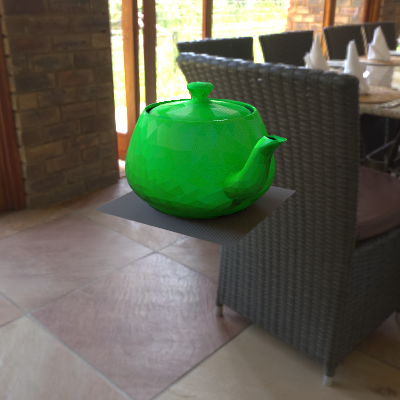}
		\caption{}
		\label{fig:teapot_pred}
	\end{subfigure}
	\begin{subfigure}[t]{0.15\textwidth}
		\includegraphics[width=\textwidth]{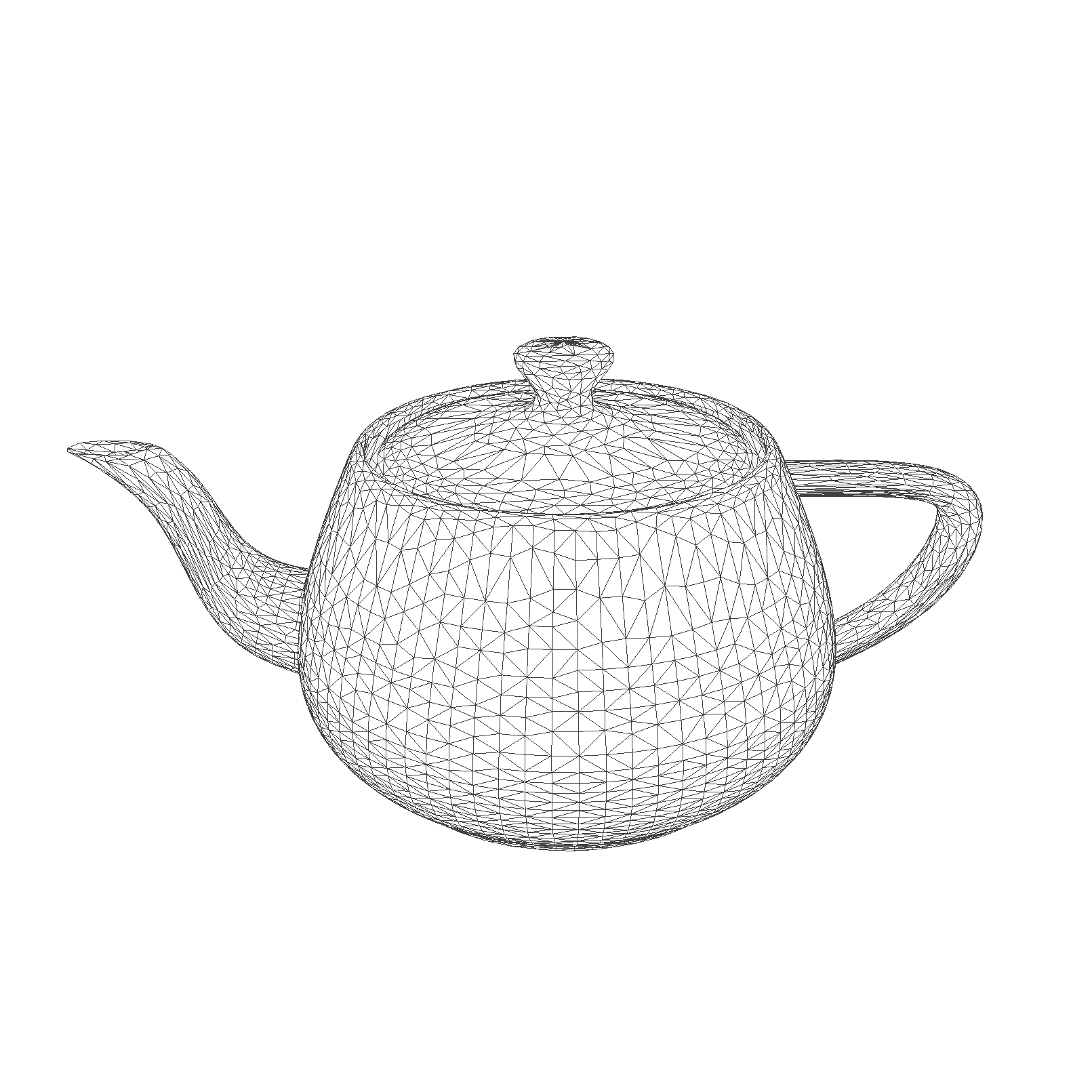}
		\caption{}
		\label{fig:teapot_mesh}
	\end{subfigure}
	
	\begin{subfigure}[t]{0.15\textwidth}
		\includegraphics[width=\textwidth]{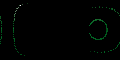}
		\caption{}
		\label{fig:map1}
	\end{subfigure}
	\begin{subfigure}[t]{0.15\textwidth}
		\includegraphics[width=\textwidth]{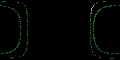}
		\caption{}
		\label{fig:map2}
	\end{subfigure}
	\begin{subfigure}[t]{0.15\textwidth}
		\includegraphics[width=\textwidth]{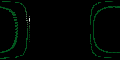}
		\caption{}
		\label{fig:map3}
	\end{subfigure}    
	\caption{Synthetic example. (a), (d): The ground truth; from these synthetic images the reflectance maps were generated as described in Section \ref{sec:method} (b),(e): A render using the predicted reflectance properties for each face under the same environment map and camera pose as (a). (c),(f): Wireframe of the mesh. Dragon model: 13,377 triangles; inference time 13.5027 seconds (1.0094ms per triangle). Teapot model: 7,809 triangles;inference time was 8.1644 seconds (1.0455ms per triangle). (g),(h),(i): reflectance maps of three randomly chosen triangles from the teapot model.} 
	\label{fig:synthetic_examples}
\end{figure}

Figure \ref{fig:synthetic_examples} presents the results on synthetic images. The synthetic images were rendered using an environment map as shown in Figure \ref{fig:dragon_gt} and were used to aggregate the outgoing radiance for each face of the dragon model. The predicted reflectance properties are then used to re-render the model under the same environment map as shown in Figure \ref{fig:dragon_pred}. The dragon model comprises of 13,377 triangles for which inference time was 13.5027 seconds (1.0094ms per triangle). Figure \ref{fig:dragon_mesh} shows the wireframe mesh. Similarly for the middle row showing the results of the Utah teapot. The model comprises of 7,809 triangles for which inference time was 8.1644 seconds (1.0455ms per triangle). We show the reflectance maps of three randomly chosen triangles from the teapot model in the bottom row. The RGBA reflectance maps are converted to RGB with a black background for legibility.

\vspace{-10pt}
\subsection{MULTI-VIEW IMAGERY}
We report results on the BlendedMVS dataset \cite{yao2020blendedmvs} and Tanks and Temples \cite{Knapitsch2017}. 

\noindent
\textbf{BlendedMVS} consists of multi-view images and their reconstructed 3D models. The reconstruction from multi-view-stereo was performed using an online platform and consists of a total of 113 scenes including cities, architectures, sculptures and small objects. Each scene contains 20 to 1,000 input images. 

\begin{figure}[!ht]
	\centering
	\begin{subfigure}[t]{0.23\textwidth}
		\includegraphics[width=\textwidth]{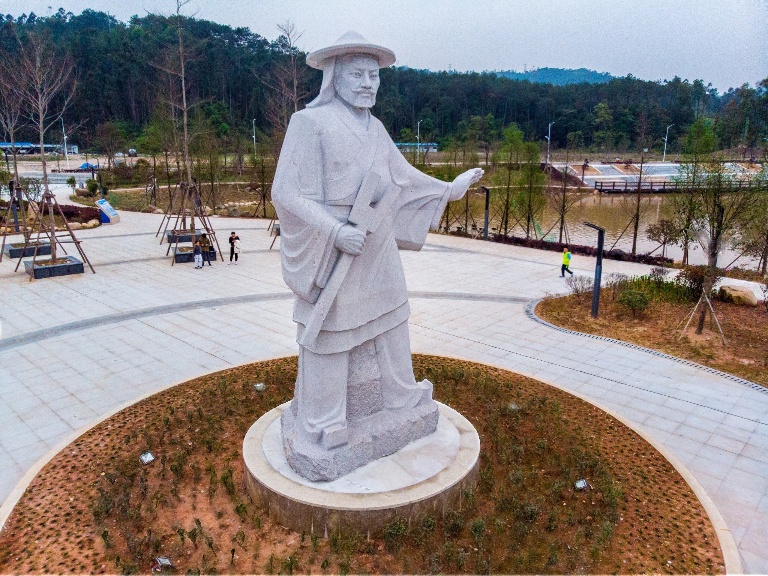}
		\caption{}
		\label{fig:statue_original1}
	\end{subfigure}
	\begin{subfigure}[t]{0.23\textwidth}
		\includegraphics[width=\textwidth]{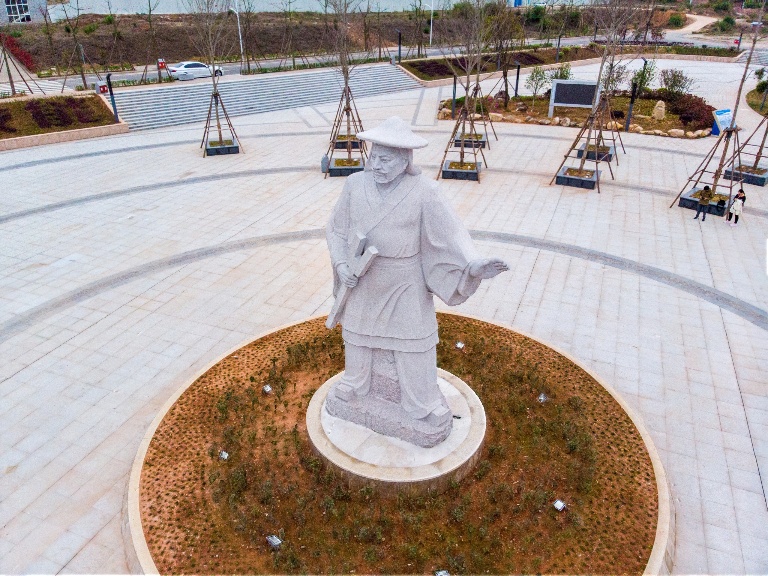}
		\caption{}
		\label{fig:statue_original2}
	\end{subfigure}
	
	\begin{subfigure}[t]{0.23\textwidth}
		\includegraphics[width=\textwidth]{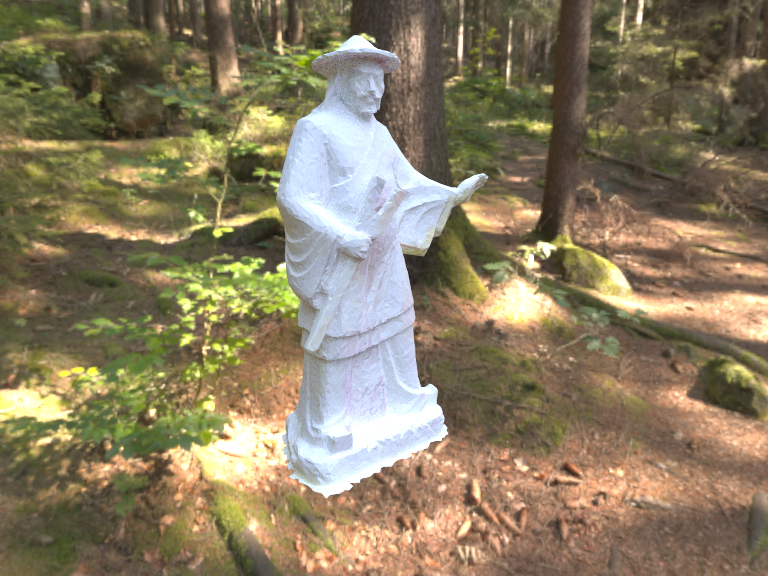}
		\caption{}
		\label{fig:statue_render1}
	\end{subfigure}
	\begin{subfigure}[t]{0.23\textwidth}
		\includegraphics[width=\textwidth]{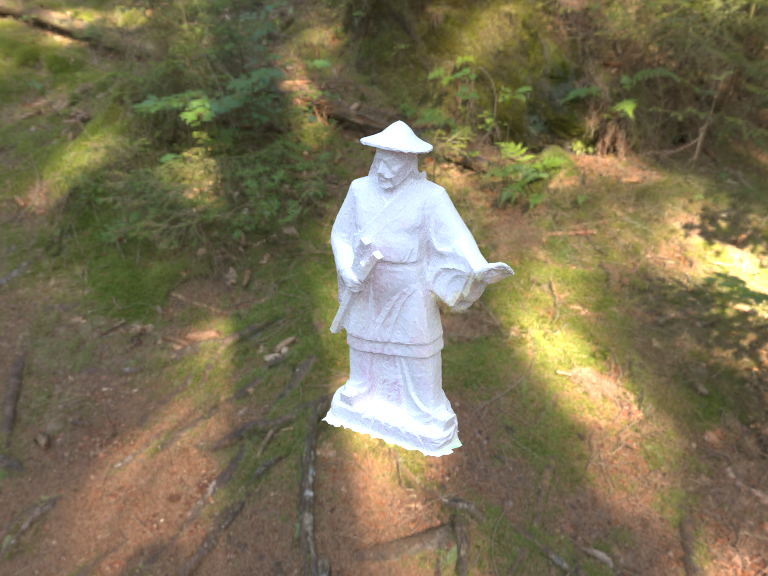}
		\caption{}
		\label{fig:statue_render2}
	\end{subfigure}
	\caption{The reflectance properties of each face of the model are predicted independently; there is no information sharing between neighbouring faces. Number of vertices/faces: 12,669/17,653; 105 images used.}
	\label{fig:statue}
\end{figure}

Figure \ref{fig:statue} shows the results of the "statue" image sequence. Reflected radiance values are aggregated into the reflectance maps from 105 images captured by an aerial camera of 768x576 pixels each. The top row shows two original images from the sequence. The second row shows the model's renders, from the same viewpoints as the originals, with the predicted per-face reflectance properties using environment mapping. The mesh consists of 17,653 faces which resulting in an equal number of aggregated reflectance maps. As explained above, the reflectance properties for each face are predicted independently without any regularization on the appearance of the neighbouring faces. Thus there is no information sharing yet the predicted reflectance properties appear visually coherent. 

\begin{figure}[!ht]
	\centering
	\begin{subfigure}[t]{0.23\textwidth}
		\includegraphics[width=\textwidth]{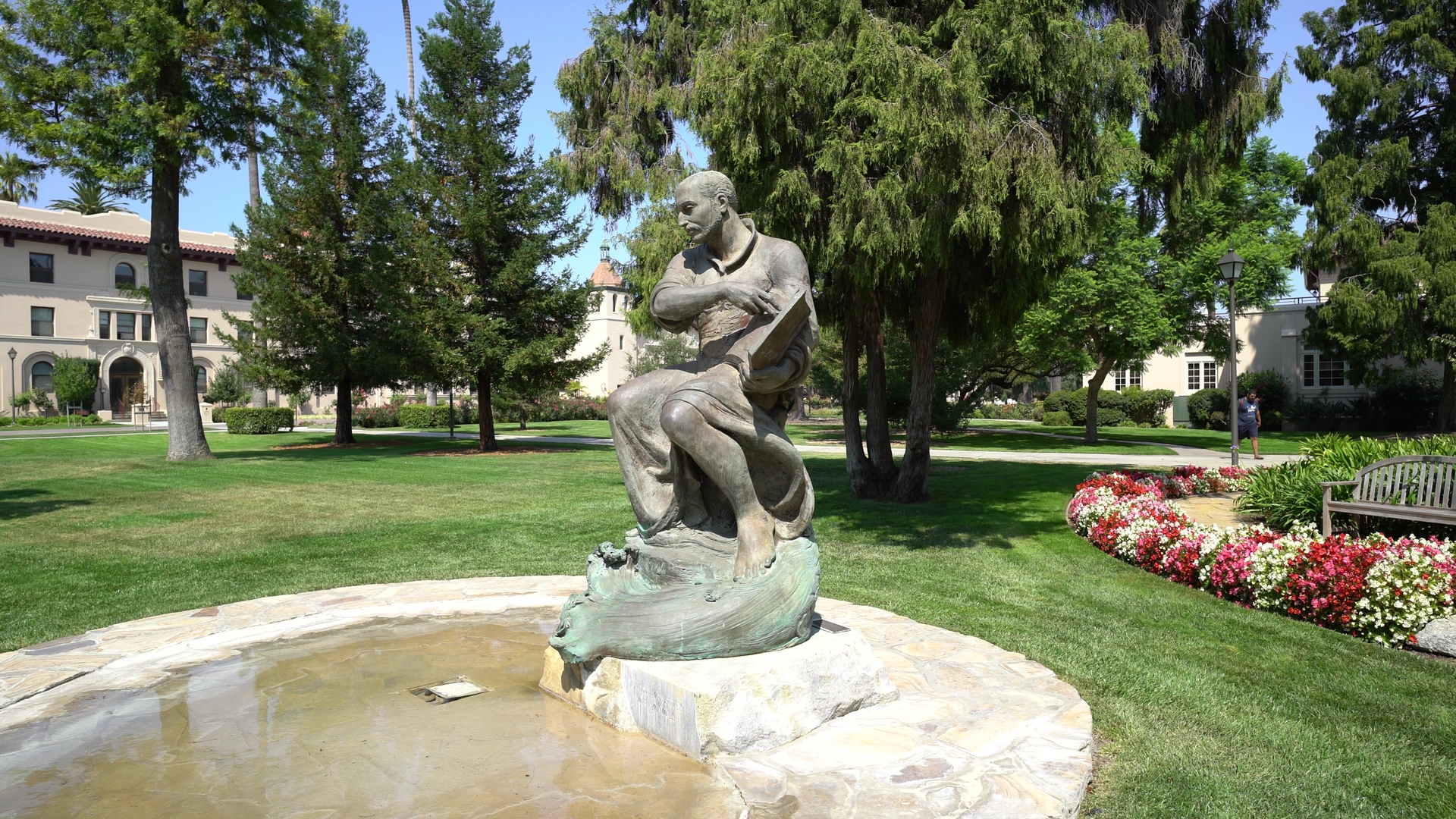}
		\caption{}
		\label{fig:ignatius_original1}
	\end{subfigure}
	\begin{subfigure}[t]{0.23\textwidth}
		\includegraphics[width=\textwidth]{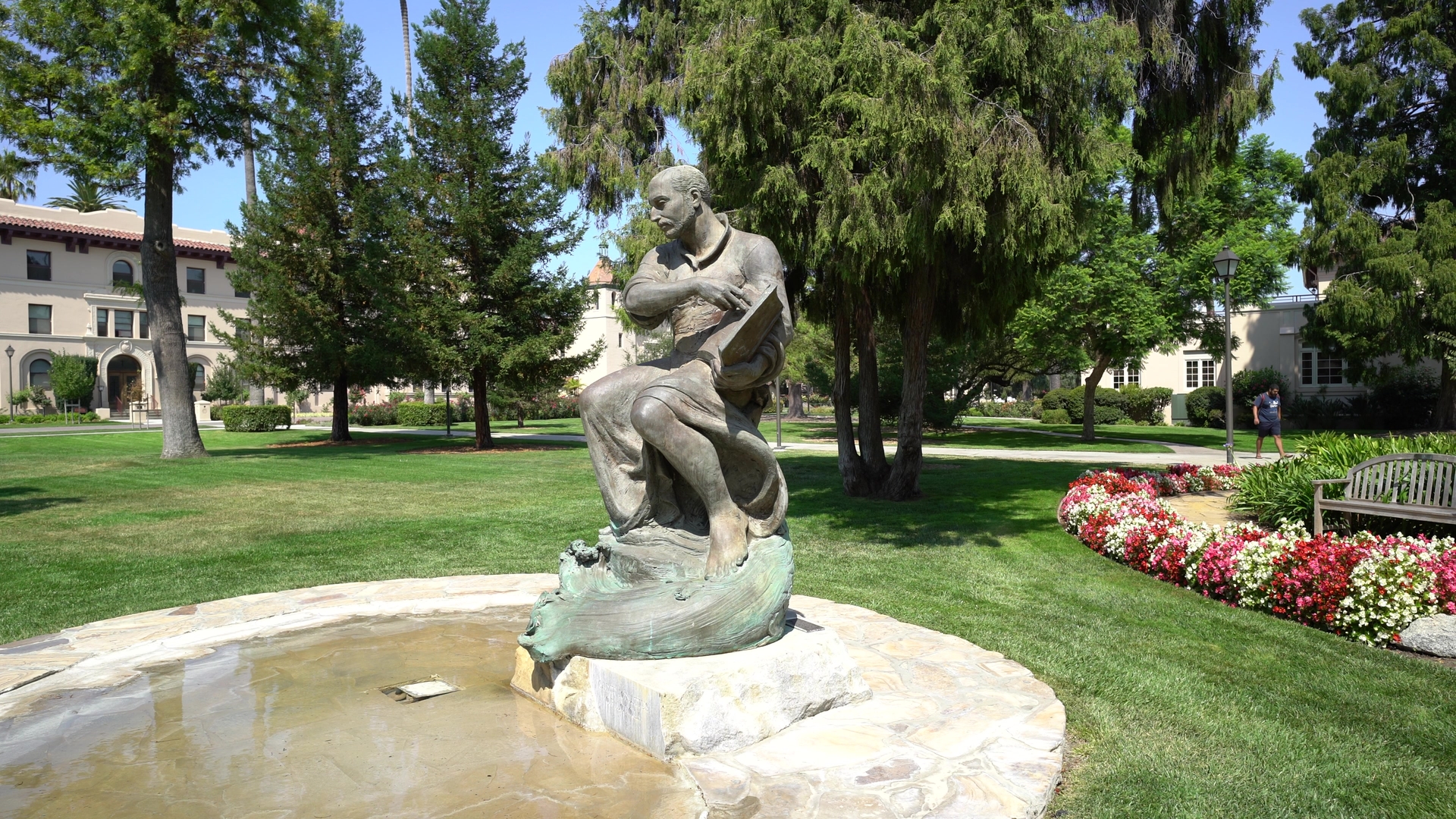}
		\caption{}
		\label{fig:ignatius_original2}
	\end{subfigure}
	
	\begin{subfigure}[t]{0.23\textwidth}
		\includegraphics[width=\textwidth]{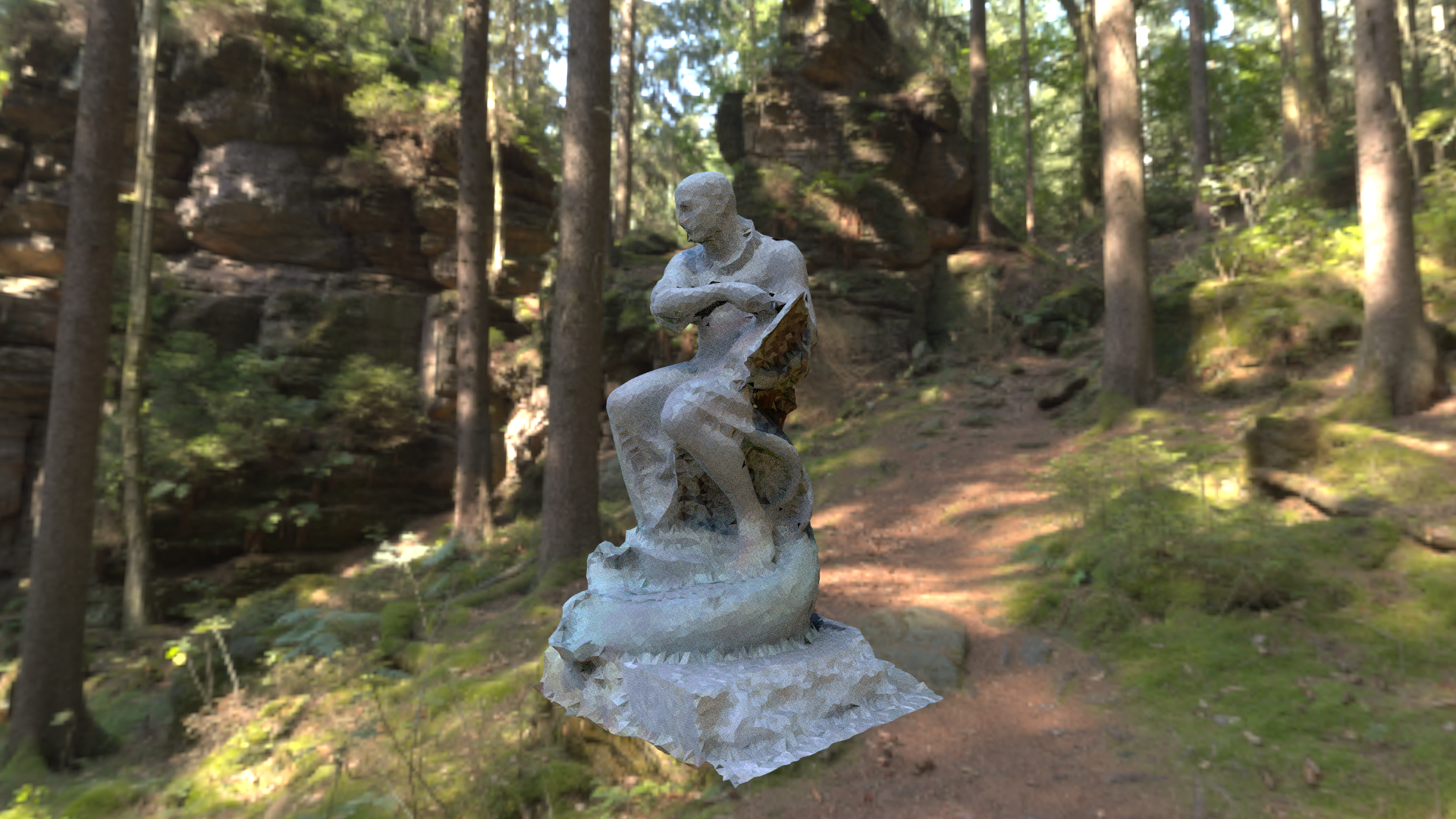}
		\caption{}
		\label{fig:ignatius_render1}
	\end{subfigure}
	\begin{subfigure}[t]{0.23\textwidth}
		\includegraphics[width=\textwidth]{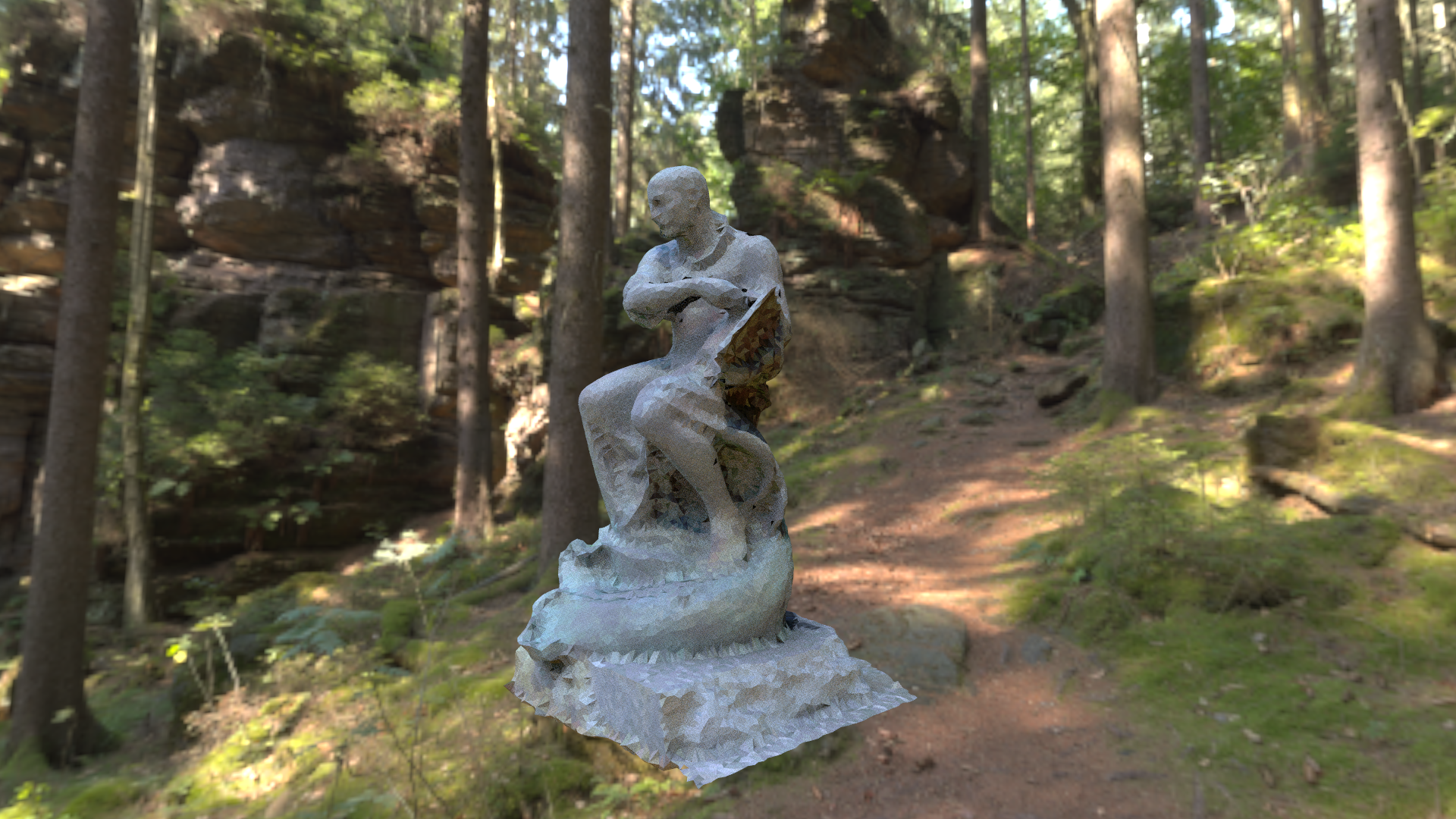}
		\caption{}
		\label{fig:ignatius_render2}
	\end{subfigure}
	\caption{Ignatius from Tanks\&Temples. The reflectance properties of each face of the model are predicted independently; there is no information sharing between neighbouring faces. Number of vertices/faces: 11,549/21,463; 31 images used.}
	\label{fig:ignatius}
\end{figure}

\begin{figure}[!ht]
	\centering
	\begin{subfigure}[t]{0.23\textwidth}
		\includegraphics[width=\textwidth]{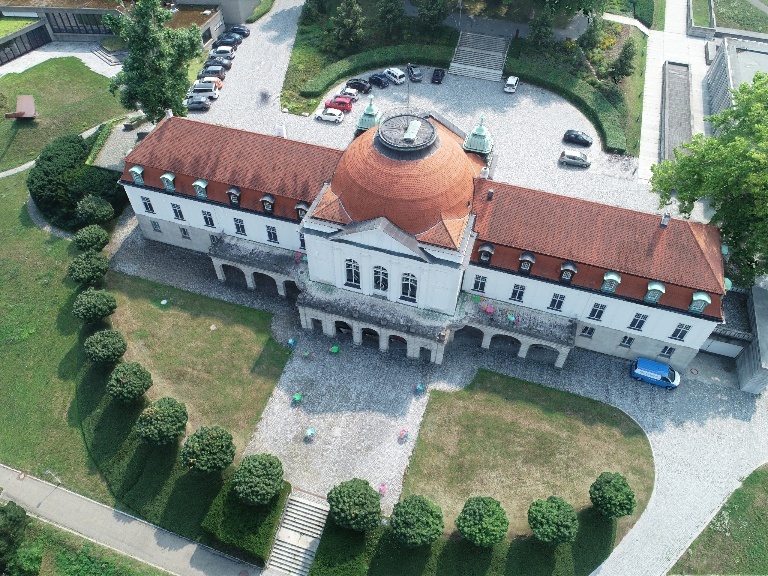}
		\caption{}
		\label{fig:building_original1}
	\end{subfigure}
	\begin{subfigure}[t]{0.23\textwidth}
		\includegraphics[width=\textwidth]{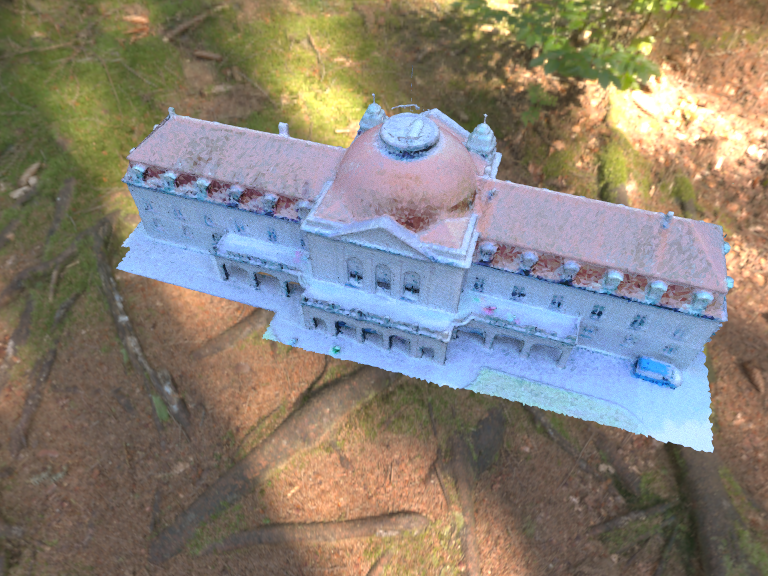}
		\caption{}
		\label{fig:building_render1}
	\end{subfigure}
	
	\begin{subfigure}[t]{0.23\textwidth}
		\includegraphics[width=\textwidth]{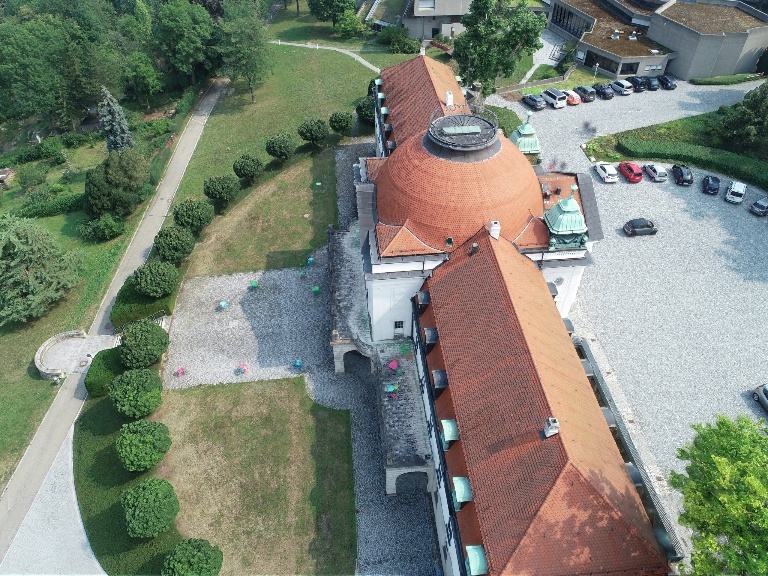}
		\caption{}
		\label{fig:building_original2}
	\end{subfigure}
	\begin{subfigure}[t]{0.23\textwidth}
		\includegraphics[width=\textwidth]{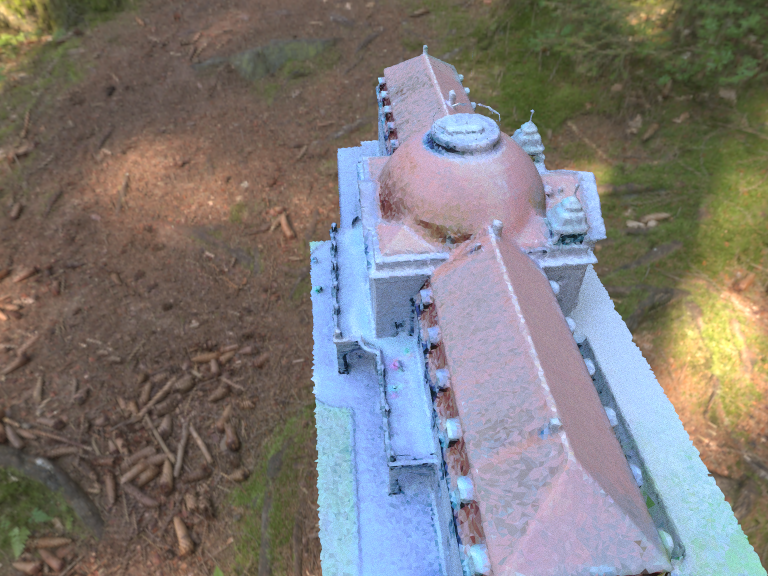}
		\caption{}
		\label{fig:building_render2}
	\end{subfigure}
	
	\begin{subfigure}[t]{0.23\textwidth}
		\includegraphics[width=\textwidth]{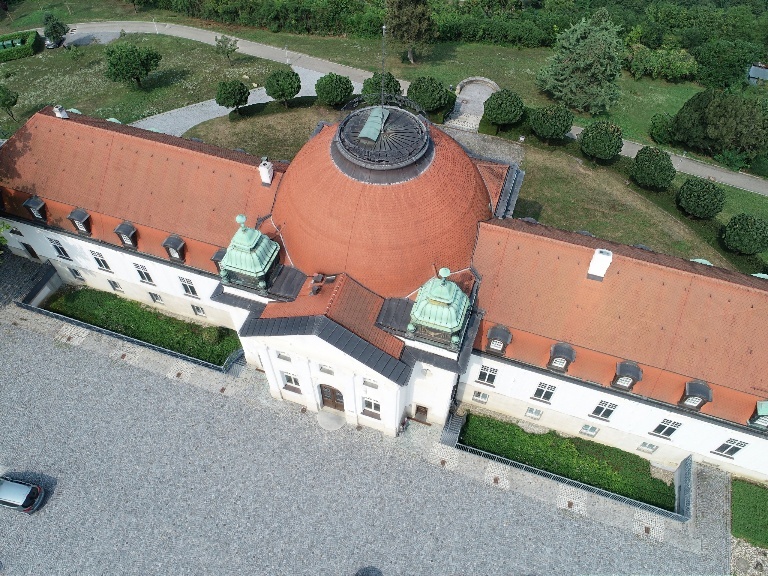}
		\caption{}
		\label{fig:building_original3}
	\end{subfigure}
	\begin{subfigure}[t]{0.23\textwidth}
		\includegraphics[width=\textwidth]{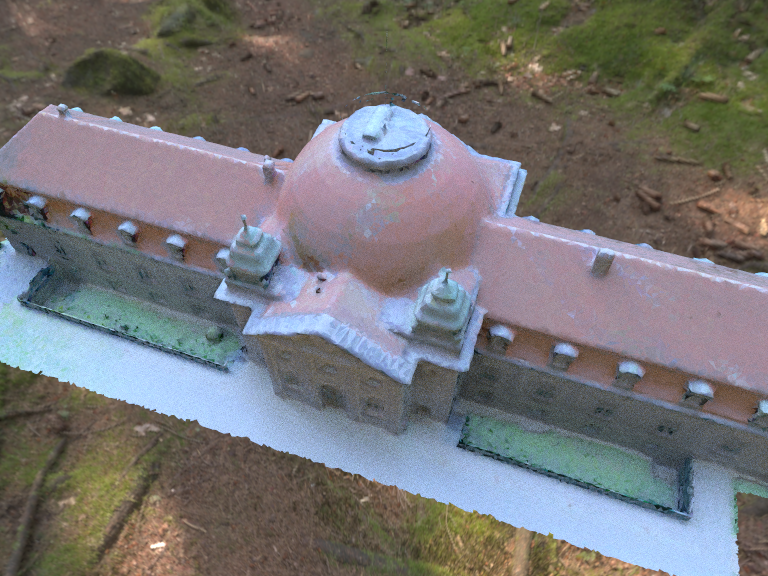}
		\caption{}
		\label{fig:building_render3}
	\end{subfigure}
	
	\begin{subfigure}[t]{0.23\textwidth}
		\includegraphics[width=\textwidth]{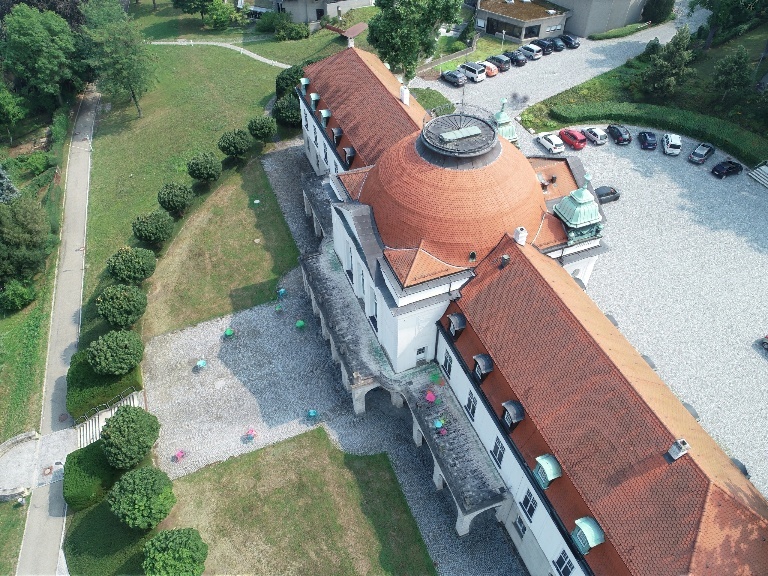}
		\caption{}
		\label{fig:building_original4}
	\end{subfigure}
	\begin{subfigure}[t]{0.23\textwidth}
		\includegraphics[width=\textwidth]{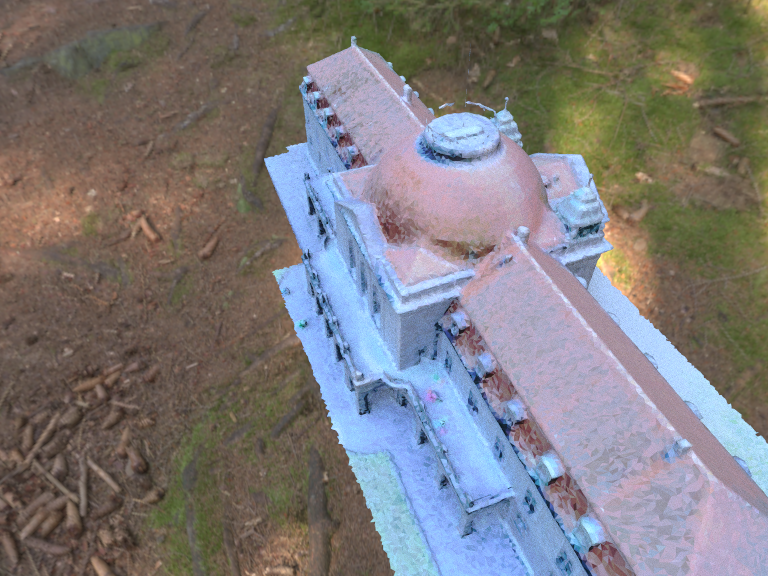}
		\caption{}
		\label{fig:building_render4}
	\end{subfigure}
	\caption{The reflectance properties of each face of the model are predicted independently; there is no information sharing between neighbouring faces. Number of vertices/faces: 50,819/100,000; 77 images used.}
	\label{fig:building}
\end{figure}

Figure \ref{fig:building} shows the results on a large building. In this experiment, we used 77 aerial images containing 768x576 pixels each, two of which are shown in the left column. The mesh consisted of 100,000 faces and 50,819 vertices. The right column shows renders using the predicted reflectance properties under directional white light. The renders using the predicted reflectance properties appear visually similar to the original when rendered. However, the proposed approach does not predict the \textit{actual} reflectance properties - because that would require knowing the texture - but instead provides an approximation that appears visually similar to the original when rendered. Moreover, due to the scale of the outdoor scenes we are considering, we assume that indirect lighting is minimal and direct lighting is from a directional light source i.e. sun; which is why the network is trained on renders of spheres being lit but a distant directional light source.

It is important to note that the level of decimation of the mesh determines the spatial resolution of the predicted BRDFs. If a low number of triangles is used then less details can be captured since each triangle is assigned one BRDF. On the other hand, a finer decimation leads to a high number of triangles and can therefore model more complex details. Thus, there is a trade-off between the number of triangles and the level of detail captured.  

\noindent
\textbf{Tanks\&Temples} provides multi-view imagery and a detailed scanned model using a sub-millimeter accuracy laser scanner. A reconstructed model using COLMAP \cite{schonberger2016structure} is also provided. Figure \ref{fig:ignatius} shows the results for the image sequence "Ignatius". The sequence consists of 31 images of 1920x1080. The bottom row shows renders of the statue using the predicted reflectance properties under an environment map. Due to the large size of the laser scanned model, we opted for using the model generated using COLMAP.


\section{CONCLUSION}
In this work, we presented a novel technique to predict reflectance properties for outdoor scenes. We have reformulated the problem of image relighting as a two-stage process following a divide and conquer approach. As part of the first stage, we create a synthetic dataset consisting of renders of spheres with various materials from different viewpoints and under arbitrary illumination. This dataset is used to compute the reflectance map and train a Wide Resnet-50 in the second stage. We justify the choice of the network architecture based on a quantitative evaluation of various architectures. Moreover, our approach can predict per-face reflectance properties in seconds when applied to models of 50,000-100,000 faces. 

\vspace{-3pt}
As part of the future work, we will focus on extending the technique to account for the effects of indirect lighting. As explained in the experimental results section, our approach predicts spatially-varying reflectance properties for the models which when used to render the model in the \textit{same environment as the original} will appear visually similar. In other words, there is no decoupling of the actual reflectance properties from the effects of indirect lighting but instead we predict reflectance properties which approximate the real reflectance properties when rendered in an environment that has the same or similar indirect lighting. 

\vspace{-3pt}
In the context of our work, the objective is to create a digital twin of the real world and therefore this is not a limitation since there is no scenario where the buildings we are considering will be used outside their original environment and context. Our objective is the realistic appearance of the objects when placed in-situ in the virtual world i.e. surrounded by a virtual environment corresponding to the real as captured in the images, under the same or similar natural illumination. Within this context the effects of indirect lighting are less significant and are encoded in the predicted reflectance properties. In our future work we would like to extend the approach to general cases where the objects can be used in a completely different environment from the one they were captured in.

\section{ACKNOWLEDGEMENTS}
This research is supported in part by the Natural Sciences and Engineering Research Council of Canada Grants DG-N01670 (Discovery Grant) and DND-N01885 (Collaborative Research and Development with the Department of National Defence Grant). 


\vspace{-10pt}
{\tiny
\bibliographystyle{IEEEtran}
\bibliography{egbib}
}

\vspace{-10pt}
\begin{IEEEbiography}{Farhan Rahman Wasee}{\,} is
currently with Clearvision, British Columbia, Canada. He received his Masters degree from Concordia University". Contact him at farhan.wasee@northsouth.edu.
\end{IEEEbiography}

\vspace{-10pt}
\begin{IEEEbiography}{Alen Joy}{\,} is currently pursuing his Masters degree at Concordia University. Contact him at alenj445@gmail.com.
\end{IEEEbiography}

\vspace{-10pt}
\begin{IEEEbiography}{Charalambos Poullis}{\,} is currently with Concordia University.  More information about his work can be found on his website www.poullis.org and on his lab's website www.theICTlab.org. Contact him/her at charalambos@poullis.org.
\end{IEEEbiography}

\end{document}